\documentclass{article}

\usepackage{microtype}
\usepackage{graphicx}
\usepackage{arydshln}
\usepackage{subfig}
\usepackage{booktabs} % for professional tables
\usepackage{amsmath,amsfonts,bm}
\usepackage{algorithm}
\usepackage{algorithmic}

\usepackage{soul}
\usepackage{xcolor}

\usepackage{hyperref}
\newcommand{\oracle}{\texttt{Oracle} }
\newcommand{\pens}{\textsc{pens} }
\newcommand{\randomsubset}{\texttt{Random} }
\newcommand{\local}{\texttt{Local} }
\newcommand{\centralnoniid}{\texttt{Central} }

\usepackage[accepted]{icml2021}

\icmltitlerunning{Decentralized federated learning of deep neural networks on non-iid data}

\begin{document}

\twocolumn[
\icmltitle{Decentralized federated learning of deep neural networks on non-iid data}



\icmlsetsymbol{equal}{*}

\begin{icmlauthorlist}
\icmlauthor{Noa Onoszko}{equal,rise,cha}
\icmlauthor{Gustav Karlsson}{equal,rise,cha}
\icmlauthor{Olof Mogren}{rise}
\icmlauthor{Edvin Listo Zec}{rise,kth}

\end{icmlauthorlist}

\icmlaffiliation{rise}{RISE Research Institutes of Sweden}
\icmlaffiliation{cha}{Chalmers University of Technology}
\icmlaffiliation{kth}{KTH Royal Institute of Technology}

\icmlcorrespondingauthor{Edvin Listo Zec}{edvin.listo.zec@ri.se}
\icmlcorrespondingauthor{Noa Onoszko}{noa.onoszko@gmail.com}

\icmlkeywords{Machine Learning, ICML}

\vskip 0.3in
]



\printAffiliationsAndNotice{\icmlEqualContribution} % otherwise use the standard text.

\begin{abstract}
    We tackle the non-convex problem of learning a personalized deep learning model in a decentralized setting. More specifically, we study decentralized federated learning, a peer-to-peer setting where data is distributed among many clients and where there is no central server to orchestrate the training. In real world scenarios, the data distributions are often heterogeneous between clients. Therefore, in this work we study the problem of how to efficiently learn a model in a peer-to-peer system with non-iid client data. We propose a method named Performance-Based Neighbor Selection (\textsc{pens}) where clients with similar data distributions detect each other and cooperate by evaluating their training losses on each other's data to learn a model suitable for the local data distribution. Our experiments on benchmark datasets show that our proposed method is able to achieve higher accuracies as compared to strong baselines.

    %The velocity at which data is being generated and gathered is still increasing worldwide. With stronger sensors on many IoT devices high quality data is gathered and stored locally, which leads to an increased demand for federated systems that harness the local computation resources. This demand only grows stronger as user privacy awareness is increasing. In traditional federated learning, a server orchestrates the learning of a global model by aggregating local models learned from client data that is never collected centrally. This global model is seldom locally optimal in real world cases where the client data is heterogeneous. We therefore in this work address the problem of data heterogeneity in peer to peer decentralized federated learning. We propose an algorithm we name Performance-Based Neighbour Selection (\textsc{Pens}{}), that effectively leverages the data heterogeneity over devices. The \textsc{Pens}{} is based on a collaborative client communication method where clients that have a similar data distribution collaborate. Our experiments show that this communication scheme results in higher model accuracies than if devices communicate randomly with each other. The method is robust for different numbers of participating devices.
\end{abstract}

\section{Introduction}
Federated learning (FL) \cite{mcmahan2017communication} is a framework developed to enable learning when data is distributed over several devices or across organizations, typically referred to as nodes or clients. In this framework, the training data never leaves the client, and all computations using the data are performed locally. This is especially useful when data privacy is important, or when collecting and storing data centrally is expensive.

Federated learning can be grouped into one of two categories: \textbf{centralized} and \textbf{decentralized}. In centralized FL, a central server orchestrates the learning among clients and is responsible for parameter aggregation, after receiving parameter updates from clients. However, the central server in an FL setup is a potential point of weakness: it could fail or be maliciously attacked, which would make the distributed learning fail. \textit{Decentralized} (peer-to-peer) systems without a central server are not vulnerable to this.

In \textit{decentralized} federated learning, no global model state exists. Instead, the participating clients follow a communication protocol to reach a consensus of a model during training. Standard techniques for decentralized learning include gradient-based algorithms based on gossip learning \cite{boyd2006randomized, jelasity2007gossip, ormandi2013gossip}, where clients train their own model based on local data and follow a communication protocol where they randomly communicate (gossip) their model parameters with their neighbors. The goal for the participating clients is to reach a consensus on a good model. In this work, we focus on gradient-based learning algorithms in a decentralized federated setup.

Both centralized and decentralized federated learning approach the important question of how to learn a suitable personalized model when client data distributions differ, i.e. the setting of non-iid data. A lot of research is currently being done regarding this topic in the centralized setting. Meanwhile, this is a relatively understudied problem in the decentralized setting \cite{kairouz2019advances}.

\textbf{The main contribution} of this paper is a novel, completely decentralized, federated algorithm for gradient-based methods when client data is non-iid: Performance-Based Neighbor Selection (\textsc{Pens}). In \textsc{Pens}, clients with similar data distributions have a higher probability of collaborating and those with dissimilar data distributions have a lower probability of collaborating. We perform multiple experiments over different non-convex optimization using deep neural networks, and our results show that using \textsc{Pens} leads to a higher performance as compared to all considered baselines.

\section{Related work}
\textbf{Gossip learning.}
Gossip learning has been applied in many different machine learning settings \cite{kempe2003gossip,boyd2006randomized,ormandi2013gossip}. However, much of the previous work on gossip learning has been limited to settings where each client only stores a single data point. Further, it has been under-explored how non-convex optimization of neural networks works under the gossip learning protocol. In \cite{giaretta2019gossip}, the authors study the performance of SVMs and linear regression models on non-iid data in a decentralized gossip learning setup. In \cite{hegedHus2019gossip} the authors train and evaluate logistic regression models and compare gossip learning to federated learning with a central server. A gossip-based algorithm for strongly convex functions has been studied in \cite{koloskova2019decentralized}, where the authors prove that their proposed algorithm is linearly convergent with quantized communication.

The first decentralized work on gossip-based optimization for non-convex deep learning studied CNNs and experimentally showed that an asynchronous and decentralized framework achieved high accuracies with low communication costs \cite{blot2016gossip}. Training CNNs in a decentralized federated learning setting has also been applied for segmentation of brain images \cite{roy2019braintorrent}.

Some recent work study communication costs (peer-to-peer communication) in non-convex optimization for different types of network topologies \cite{assran2019stochastic,wang2019matcha}.

In \cite{kong2021consensus}, the authors identify the changing consensus distance between clients as key to explain the gap between centralized and decentralized training and focus on non-convex optimization.

\textbf{Non-iid data.}
All aforementioned works solve important problems. However, a key assumption is made in the studies: that data is independently and identically distributed (iid) over clients. The problem of non-iid is becoming more studied in the case of centralized federated learning. In this setting, solutions for skewed data distributions have been explored in many different ways including fine-tuning a global model locally \cite{wang2019federated}, posing the personalization problem in FL as a meta-learning objective \cite{jiang2019improving}, using knowledge distillation techniques \cite{jeong2018communication}, by mixing local and global models \cite{deng2020adaptive,listozec2020federated} and with data-sharing methods \cite{zhao2018federated}. Meanwhile, similar techniques have not yet been widely applied and researched in decentralized federated learning.

In \cite{ghosh2020efficient} they study the problem of covariate shift, similar to us, but in a centralized federated learning setup. In their paper, the authors develop a client clustering framework to learn one global model per cluster with a central parameter server.

Label distribution shift for decentralized deep learning has been studied in \cite{niwa2020edge}. In this work, the authors propose to solve a dual problem that seeks to minimize a linearly constrained cost function. By solving a constrained optimization problem, their method achieves similar models among clients in the non-iid data setting.

In this work, we continue the research on the effectiveness of deep neural networks in decentralized peer-to-peer networks where data is non-iid. More specifically, we study the problem of covariate shift where $\mathcal{D}_i(x)$ varies, but the conditional distributions $\mathcal{D}_i(y|x) = \mathcal{D}_j(y|x)$ for all clients $i\neq j$.

\section{Problem formulation}
We formulate the problem as an empirical risk minimization (ERM) problem, as commonly used in statistical learning setups. The goal is to learn weights $w$ for a model by optimizing some loss over data. In a decentralized setting, we have $k$ clients that are able to communicate with neighboring clients in a communication network. We assume that each client $i$ has a data distribution $\mathcal{D}_i(x,y)$ over input features $x$ and labels $y$.

Let $\ell(w;z) : \mathbb{R}^d \to \mathbb{R}$ be the loss as a function of the model parameters $w$ and data points $z = (x,y)$. Thus, the aim of the optimization is to minimize
\begin{equation}
 \mathcal{L}_i(w) := \mathbb{E}_{z\sim\mathcal{D}_i}[\ell(w;z)] \quad \forall i = 1,\dots,k   
\end{equation}

In this work, we study the problem of covariate-shift. To do this, we create different distributions $\mathcal{D}^r(x)$ for each image dataset by rotating the images with $r$ degrees. $\mathcal{D}^0(x)$ is defined as a dataset where images have been rotated with 0 degrees of rotation, and $\mathcal{D}^{180}(x)$ with 180 degrees of rotation. 

We perform experiments on two and four different data distributions, where $r\in\{0,180\}$ or $r\in \{0,90,180,270\}$. The train and test sets for the studied datasets are randomly split into one equally large partition for each value of $r$, and a rotation of $r$ is applied on each partition thus creating $\mathcal{D}^r(x)$. Then each client is populated with training samples uniformly from one such data distribution. Since the labels are unchanged after rotation, this means that the marginal distributions over input features differ between these groups, but the conditional distributions $\mathcal{D}(y|x)$ are the same for all clients. This way of creating different client distributions has previously been used in \cite{ghosh2020efficient} for centralized federated learning.

The main challenge of this paper is that we assume the client distributions $\mathcal{D}_i^r$ are \textit{unknown for each client $i$}, and our goal is to design an algorithm that can both identify $\mathcal{D}^r_i$ and at the same time perform distributed optimization.

\section{Algorithm}
In our decentralized peer-to-peer network, we use the gossip protocol for communication between clients. Below we describe the random gossip baseline, and our proposed extension \textsc{pens}.

\subsection{Random gossip communication}
\label{gossip}
In this framework, each client starts with a randomly initialized model that is updated using stochastic gradient descent (SGD) on the local client data for $E$ local epochs. The model parameters $w_i$ of client $i$ are then at a random time communicated to a randomly chosen neighboring client $j$ in the network. This action is denoted as \textbf{Send$_{i\to j}(\cdot)$}. Client $j$ then waits for $n_{\text{peers}}$ number of models before it aggregates its own current model with the received ones with a simple average: $\bar{w}_j^{t+1} \leftarrow \frac{1}{n}\sum_i w_i^t$. This is the same aggregation method as commonly used in centralized federated learning. The new aggregated model $\bar{w}_j^{t+1}$ is then trained for $E$ epochs before it is ready to be gossiped again. A summary of the gossip learning protocol is presented in algorithm \ref{gossip_alg}.

\subsection{\textsc{Pens}: Performance-based neighbor selection}
A problem that arises with random gossip when we have different distributions $\mathcal{D}^r_i(x)$ for clients $i$ is that if two clients with dissimilar distributions (i.e. $r=0$ for client $i$ and $r=180$ for client $j$) communicate, the performance of the learned model is usually negatively effected when their models are aggregated. 

To solve this problem, we introduce \textsc{Pens}: Performance-based neighbor selection. \textsc{Pens} consists of two main steps. In the first step, the algorithm finds clients of similar marginal distributions $\mathcal{D}_i(x)$ to communicate with. In the second step, the random gossip protocol is followed for the subset of clients selected from the first step. \textit{Clients of similar distributions are found by evaluating sent models on the receiving client's training data.} The main idea of the proposed method is that the training loss of a sent client model $w_i$ is expected to be lower on the training set of a receiving client $j$ that has a similar data distribution, and a higher loss for those clients that have dissimilar distributions.

First, each client communicates randomly in the network for a pre-defined number of neighbor selection communication rounds $T$. At a random time, a client performs the \textbf{Send$_{i\to j}(\cdot)$} operation, after which $\mathcal{L}_j(w_i ; z)$ is calculated, where $z\sim \mathcal{D}_j$. This is the loss of client model $w_i$ on the training set of client $j$. Each client waits for $n_{\text{sampled}}$ number of models and saves a list of the losses. The top $m$ best performing (lowest loss) clients are selected as \textit{potential neighbors with similar data distributions}. Then their model parameters are aggregated into a new model. This is repeated for $T$ rounds, after which the clients that were selected more than the expected amount of times (if the sampling of clients would have been uniform) are identified as \textit{neighbors with similar data distributions}.

A set of neighbors with similar data distributions are now identified for every client and this constitutes step 1 of \textsc{Pens}. This is summarized in algorithm \ref{pens_alg}. In step 2 of \textsc{Pens}, the gossip learning protocol (algorithm \ref{gossip_alg}) is used for the set of selected neighbors for each client.

\begin{algorithm}[h!]
  \caption{Gossip learning protocol}
  \label{gossip_alg}
  \begin{algorithmic}[1] % The number tells where the line numbering should start
    \FUNCTION{\texttt{MAIN}}
        \WHILE{stopping criterion not met}
            \STATE \texttt{WAIT($\Delta$)}
            \STATE $j \gets$ \texttt{RANDOMPEER()} \textit{// select random peer}
            \STATE \texttt{SEND$_{i\to j}$($w_i,j$)}
        \ENDWHILE
    \ENDFUNCTION
    \FUNCTION{\texttt{ONRECEIVEMODEL}($w_i$)}
    \STATE \texttt{SAVE}($w_i$)
        \IF{no. of received models $\geq n_{\text{peers}}$}
           \STATE $w_j \gets $\texttt{MERGE\_SAVED\_MODELS}()
            \STATE $w_j \gets $\texttt{TRAIN}($x;w_j$) //\textit{update on local data $x$}
        \ENDIF
    \ENDFUNCTION
  \end{algorithmic}
\end{algorithm}

\begin{algorithm}[h!]
  \caption{\texttt{PENS} step 1: find peers with similar data distributions $\mathcal{D}^r_i(x)$}
  \label{pens_alg}
  \begin{algorithmic}[1]
    \FUNCTION{\texttt{MAIN}}
        \WHILE{stopping criterion not met}
            \STATE \texttt{WAIT($\Delta$)}
            \STATE $j \gets$ \texttt{RANDOMPEER()} \textit{// select random peer}
            \STATE \texttt{SEND$_{i\to j}$($w_i,j$)}
        \ENDWHILE
    \STATE \texttt{SELECTNEIGHBORS}()
    \ENDFUNCTION
    \FUNCTION{\texttt{ONRECEIVEMODEL($w_i$)}}
        \STATE $\ell_i \gets$ \texttt{CALCULATELOSS($w_i$)}
        \STATE \texttt{SAVE}($w_i, \ell_i$)
        \IF{no. of received models $\geq n_{\text{sampled}}$}
            \STATE $w_j \gets $\texttt{MERGE(SELECT\_TOP\_M($w_i$,$\ell_i$))}
            \STATE $w_j \gets $\texttt{TRAIN($x;w_j$)} \textit{//update on local data $x$}
            \STATE no. of received models $\gets 0$ \textit{//reset}
        \ENDIF
    \ENDFUNCTION
    \FUNCTION{\texttt{SELECTNEIGHBORS}}
        \FORALL{peers $i$}
            \IF{merged with $i$ more than expected}
                \STATE \texttt{NEIGHBORLIST.append}($i$)
            \ENDIF
        \ENDFOR
        \STATE \textbf{return:} \texttt{NEIGHBORLIST}
    \ENDFUNCTION
  \end{algorithmic}
\end{algorithm}

\section{Experimental setup}
In this work, we set out to develop an algorithm to solve the problem of non-iid data for gradient-based algorithms in a peer-to-peer network. To do this, we limit the experiments to a peer-to-peer network that is fully connected (i.e. all nodes in the network can communicate with each other). Further, we assume that all clients are able to communicate at any time. We simulated the peer-to-peer network on a computer, and all experiments were performed with a NVIDIA Tesla V100-SXM2-32GB GPU. Our code is available at github \footnote{\url{https://github.com/guskarls/dfl-pens}}. 
\subsection{Datasets}
Our experiments are performed on two datasets for visual classification, CIFAR-10  \cite{krizhevsky2009learning} and  Fashion-MNIST \cite{xiao2017fashion}. The CIFAR-10 dataset consists of 60 000 32x32 color images in 10 classes, with 6000 images per class. The dataset is split into 50 000 training images and 10 000 test images. The Fashion-MNIST dataset contains 70 000 28x28 gray-scale images of Zalando clothing in 10 classes. It is split into 60 000 training images and 10 000 test images.

\subsection{Models and hyperparameters}
The CNN used in our experiments consists of three convolutional layers and ReLU activations (with 32 channels in the first layer and 64 channels in the last two layers), each with a kernel size of $3$ and each followed by max pooling. This is followed by one fully connected layer of $64$ units with a ReLU activation and an output layer with a softmax activation. The size and architecture of this network is not state-of-the-art for visual classification tasks but has sufficient capacity for the comparison that we perform in our experiments. We use SGD as our optimizer with a learning rate $\eta=10^{-3}$ for \textsc{pens}. For the random gossip protocol we use $n_{\text{peers}} = 20$ and for \pens $n_{\text{sampled}} = 10$ and $m=2$, if not explicitly otherwise stated. All hyperparameters were tuned for all baselines and the best ones were chosen with respect to a local validation set on each client.

\subsection{Baselines}
We compare our proposed algorithm \textsc{Pens} with two baselines: random gossip (\texttt{Random}) and locally trained models without communication (\texttt{Local}) for each client. We also report results for an \texttt{Oracle}, that for each client is given the information of which neighbors have the same distributions $\mathcal{D}^r_i(x)$ for all $i=1,\dots,k$, and only communicates with these neighbors. Accuracy for a centrally trained model is also presented, denoted by \texttt{Central}, where we train one model on all data non-distributed.

\subsection{Evaluation}
For testing of all algorithms, we measure test accuracy for each client $i$ on test data from the client's own distribution $\mathcal{D}^r_i(x)$. All reported accuracies are averaged over clients for each distribution. We run 4 experiments for each algorithm, with different random seeds, and report the average and a $95\%$ confidence interval. For step 1 of \textsc{pens}, we let clients communicate $T=200$ communication rounds. During step 2 of \textsc{pens}, \texttt{Random} and \texttt{Oracle}, we set $T=333$ and we perform early stopping on local validation data for each client between client communication rounds. The communication is stopped when the validation loss has converged or when we reach $T$. The validation sets consist of 100 sample points from $\mathcal{D}^r_i$ per client in all experiments.

\section{Results and discussion}
In table \ref{tab:cifar10}, results on CIFAR-10 are shown for $r=\{0,180\}$. Accuracies are reported both for independent and common weight initialization for the client models. In centralized federated learning, it is known that a common initialization of client models is important for federated averaging to work \cite{mcmahan2017communication}. Meanwhile, our results suggest that a common initialization is not necessary in order for the different algorithms to reach a high accuracy in decentralized federated learning. Further, the proposed method \pens achieves an accuracy that is higher than all baselines and close to the performance of \oracle with perfect information of client data distributions $\mathcal{D}^r_i(x)$. In table \ref{tab:4cluster} we see that \pens outperforms the baselines also in the case of $r=\{0,90,180,270\}$.

In table \ref{tab:fashion}, accuracies for all algorithms on Fashion-MNIST with $r=\{0,180\}$ are presented for 100 and 500 training samples per client, with the number of clients set to 100. Although the difference of test accuracy between the baselines is smaller as compared to CIFAR-10 (since Fashion-MNIST is an easier problem), our proposed method \pens outperforms both baselines in this setting as well.
\begin{figure*}[t]
    \centering
    \subfloat[Number of clients fixed to 100. Training set size per client varying.]{\label{datapoints}\includegraphics[width=.45\textwidth]{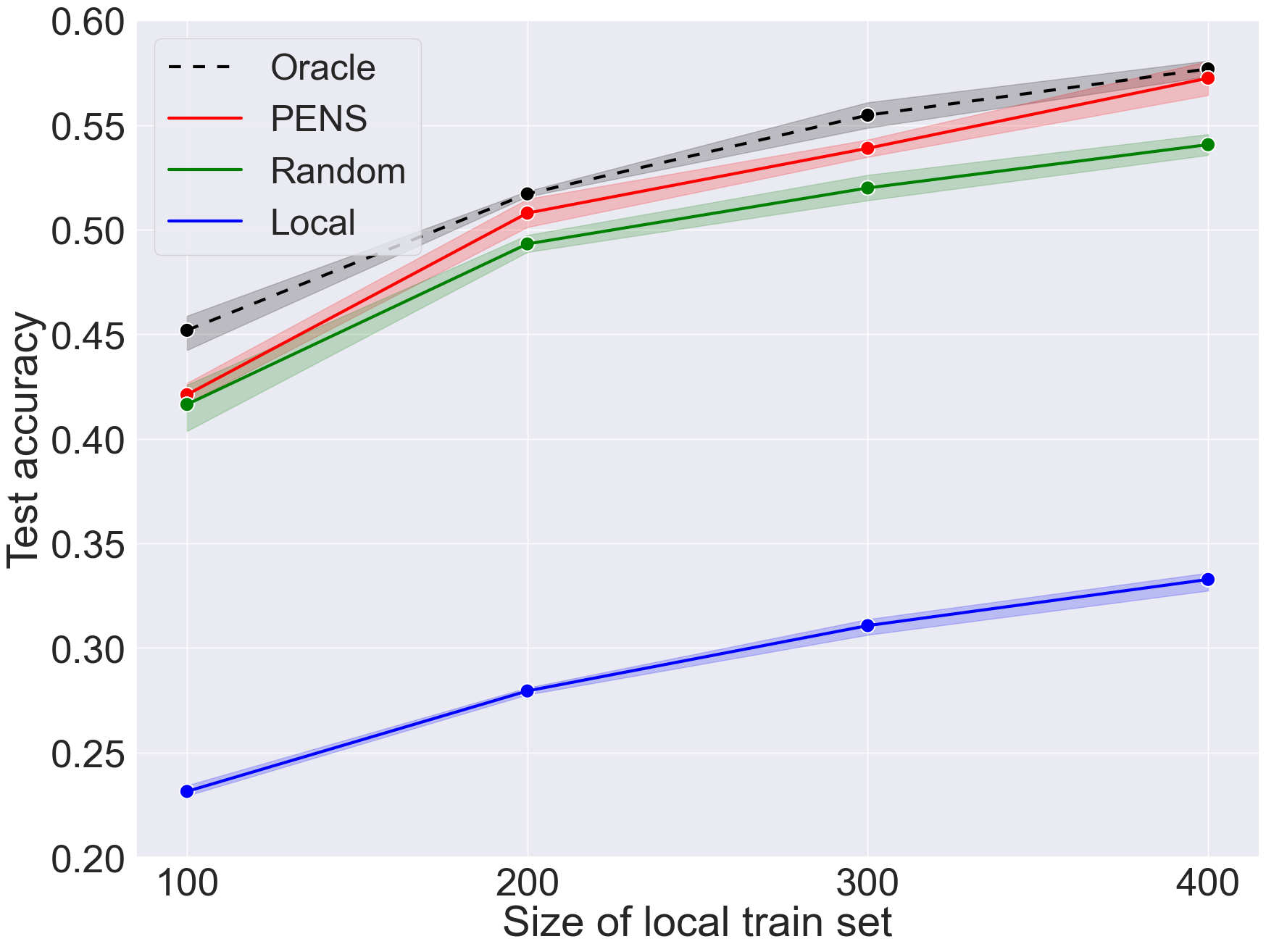}}\hfill
    \subfloat[Train set size per client fixed to 150. Number of clients varying.]{\label{clients}\includegraphics[width=.45\textwidth]{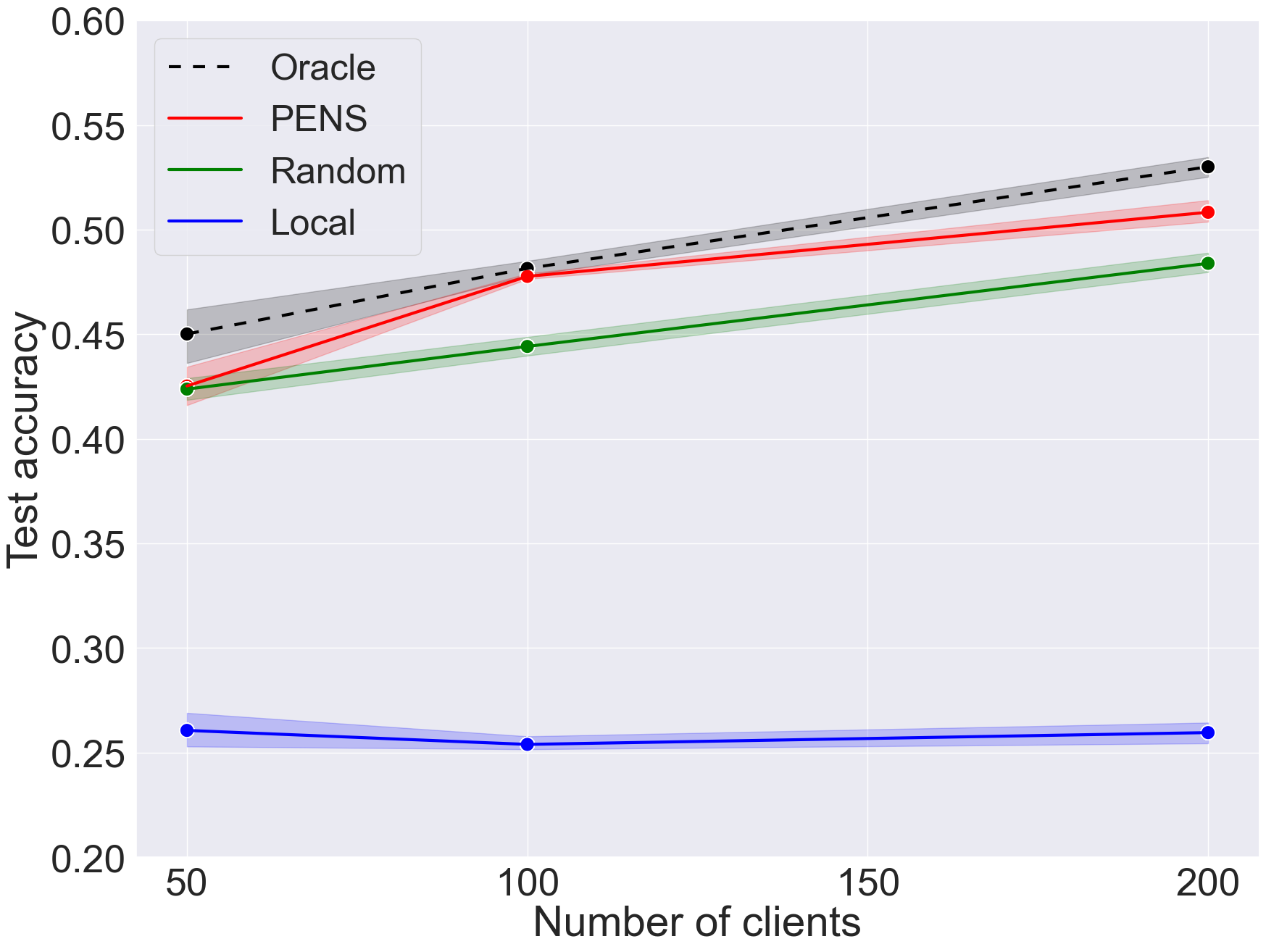}}
    \caption{Test accuracy on CIFAR-10 with $r=\{0,180\}$ as a function of (a) training samples per client and (b) number of clients, while fixing the other. \oracle has perfect information of client distributions, as opposed to the other methods.}
    \label{datapoints_clients}
\end{figure*}
\subsection{Impact of training set size}
In figure \ref{datapoints} we compare \pens to the baseline algorithms on the CIFAR-10 dataset in a setting where we fix the number of clients to 100 while at the same time varying the size of the local train sets. The results show that by increasing the size of the local train set on each client, performance increases for all compared algorithms. We further see that \pens consistently outperforms both baselines. In a low data setting, with 100 training samples per client, \pens is closer to \randomsubset in performance as compared to \texttt{Oracle}. However, when the training size increases, the difference between \randomsubset and \pens increases, as \pens manages to find the correct neighbors with similar data distributions for each client. This is further visualized in figure \ref{heatmaps}, where a heatmap over the communication pattern is plotted for \texttt{Oracle}, \textsc{pens}, and \texttt{Random}. Here we see that for each client, \pens manages to find almost all clients with similar data distributions.

In table \ref{tab:precision} we report for precision (the fraction of clients with the same distribution $\mathcal{D}^r$ among the selected clients) and recall (the fraction of clients with the same distribution $\mathcal{D}^r$) for the peers that \pens{} selects for each client. Here we report experiments for 100 clients for a different number of training samples per client. We note that the precision of our method is very high and robust to the number of training samples per client. The recall is lower than the precision, but also robust to the number of training samples.

\subsection{Impact of number of clients}
Figure \ref{clients} shows results for experiments for CIFAR-10 where the number of samples per client is fixed to 250 (150 train and 100 validation samples), but with a varying number of clients in the peer-to-peer network. Since no communication is allowed for the local baseline, the performance is constant with respect to the number of clients. Meanwhile, for the other methods, we see that by adding more clients (and therefore also the total amount of data in the system) the performance increases. Further, our proposed method \pens consistently outperforms the random baseline. 
    \begin{table}[h!]
        \centering
        \caption{Test accuracy reported on CIFAR-10. Independent and common model weight initialization. 100 clients, 400 training samples.}
        \begin{tabular}{c|c|c}
            Method & Acc. (independent) \(\%\)&  Acc. (common)  \(\%\) \\ \hline
            \textcolor{gray}{Central} & \textcolor{gray}{$65.5\pm0.3$} & \textcolor{gray}{$65.5\pm0.3$} \\ 
            \textcolor{gray}{Oracle} & \textcolor{gray}{$58.1\pm0.4$} & \textcolor{gray}{$57.7\pm0.6$}\\ \hdashline
            \pens & $\mathbf{57.8\pm0.4}$ &  $\mathbf{57.2\pm1.3}$\\
            Random & $54.7\pm0.8$ & $54.1\pm0.8$ \\
            Local & $33.4\pm0.2$ & $33.3\pm0.7$ \\
        \end{tabular}
        \label{tab:cifar10}
    \end{table}
\subsection{Impact of $n_{\text{sampled}}$ and top $m$ performers}
The parameter $n_{\text{sampled}}$ in step 1 of \pens decides for each client how many other client models to sample at every communication round, and $m$ decides how many of the top-performing (lowest loss) to merge with. Experiments were carried out to study how sensitive \pens{} is to the choice of these hyperparameters. In table \ref{tab:mn} we summarize the results for varying values of these hyperparameters in the setting of 100 clients and 400 training samples per client for CIFAR-10. We note that the test accuracies are relatively stable for different values of $n_{\text{sampled}}$ and $m$. Further, our results suggest that the ratio $\frac{n_{\text{sampled}}}{m}$ should not be too large, i.e. if $n_{\text{sampled}}$ is increased $m$ should be higher as well. We have noticed in our experiments that if $m$ is set too low relative to $n_{\text{sampled}}$, \pens will collapse into always choosing the same few clients and miss to find other peers of the same distribution.
    
    \begin{table}[h!]
        \centering
        \caption{Test accuracies on CIFAR-10 with 4 rotated distributions $\mathcal{D}^r(x)$, $r= \{0,90,180,270\}$. 100 clients and 400 training samples per client.}
        \begin{tabular}{c|c}
            Method & Accuracy (\%) \\ \hline
            \textcolor{gray}{\centralnoniid} &  \textcolor{gray}{$60.3\pm0.8$} \\
            \textcolor{gray}{\oracle} & \textcolor{gray}{$53.2\pm1.1$} \\ \hdashline
            \pens & $\mathbf{53.4\pm0.7}$ \\
            \randomsubset & $49.3\pm1.0$ \\
            \local & $33.4\pm0.7$ \\
        \end{tabular}
        \label{tab:4cluster}
    \end{table}
    
    \begin{table}[h!]
        \centering
        \caption{Test accuracy reported on Fashion-MNIST for 100 clients with 100 and 500 training samples per client and $r=\{0,180\}$.}
        \begin{tabular}{c|c|c}
            Method & 100 & 500 \\ \hline
            \textcolor{gray}{\centralnoniid} &  \textcolor{gray}{$85.3\pm0.3$} & \textcolor{gray}{$87.3\pm0.3$} \\
            \textcolor{gray}{\oracle} & \textcolor{gray}{$79.4\pm0.4$} & \textcolor{gray}{$84.9\pm0.2$} \\ \hdashline
            \pens & $\mathbf{78.8\pm0.4}$ & $\mathbf{84.7\pm0.3}$ \\
            \randomsubset & $77.5\pm0.5$ & $83.7\pm0.2$ \\
            \local & $63.9\pm0.6$  & $74.4\pm0.5$ \\
        \end{tabular}
        \label{tab:fashion}
    \end{table}

    \begin{table}[h!]
        \centering
        \caption{Accuracies for varying $n_\text{sampled}$ and $m_\text{selected}$ on CIFAR-10 for 100 clients with 150 training samples per client, $r=\{0,180\}$.}
        \begin{tabular}{c|c|c|c}
            $n_\text{sampled}$ & $m$ & $n_\text{sampled}/m$ & Accuracy (\%)\\\hline
            20 & 6 & 3.3 & $48.4\pm0.5$\\
            50 & 15 & 3.3 & $47.8\pm1.0$\\
            10 & 2 & 5.0 & $47.7\pm1.4$\\
            10 & 3 & 3.3 & $47.2\pm1.1$\\
            50 & 10 & 5.0 & $46.9\pm1.4$\\
            20 & 4 & 5.0 & $46.9\pm0.8$\\
            10 & 5 & 2.0 & $46.4\pm0.9$\\
            5 & 2 & 2.5 & $46.2\pm0.8$\\
            10 & 1 & 10.0 & $46.1\pm0.7$\\
            5 & 1 & 5.0 & $45.4\pm0.7$\\
            20 & 10 & 2.0 & $45.4\pm1.8$\\
            20 & 2 & 10.0 & $45.4\pm2.0$\\
            5 & 3 & 1.7 & $45.2\pm2.0$\\
            50 & 25 & 2.0 & $45.1\pm2.5$\\
            50 & 5 & 10.0 & $44.5\pm2.0$\\
        \end{tabular}
        \label{tab:mn}
    \end{table}   
    
    \begin{table}[h!]
        \centering
        \caption{Precision and recall for 100 clients for different number of training samples and $r=\{0,180\}$.}
        \begin{tabular}{c|c|c}
            Train set size & Precision (\%) & Recall (\%)\\
            400 & $99.7\pm0.1$ & $70.8\pm0.8$\\
            300 & $99.7\pm0.1$ & $71.1\pm0.9$\\
            200 & $99.7\pm0.1$ & $70.1\pm0.7$\\
            100 & $95.8\pm1.5$ & $67.9\pm1.5$\\
        \end{tabular}
        \label{tab:precision}
    \end{table}

    % Oracle---CIFAR-10400 comcost=$5039\pm202$
    % Oracle---Fashion-MNIST100 comcost=$4708\pm313$
    % Oracle---Fashion-MNIST500 comcost=$4295\pm170$
    % PENS---CIFAR-10400 comcost=$4848\pm98$
    % PENS---Fashion-MNIST100 comcost=$4607\pm152$
    % PENS---Fashion-MNIST500 comcost=$4253\pm211$
    % Random---CIFAR-10400 comcost=$5048\pm109$
    % Random---Fashion-MNIST100 comcost=$4637\pm335$
    % Random---Fashion-MNIST500 comcost=$4555\pm249$

% PENS---Fashion-MNIST---500: 84.7\pm0.3
% Oracle---Fashion-MNIST---100: 79.4\pm0.4
% PENS---CIFAR-10---400: 57.8\pm0.4
% Random---CIFAR-10---400: 54.7\pm0.8
% Random---Fashion-MNIST---500: 83.7\pm0.2
% Oracle---CIFAR-10---400: 58.1\pm0.4
% Random---Fashion-MNIST---100: 77.5\pm0.5
% PENS---Fashion-MNIST---100: 78.8\pm0.4
% Oracle---Fashion-MNIST---500: 84.9\pm0.2

% Central/local
% Fashion-MNIST, n_train=500, n_clients=100: 74.4\pm0.5
% CIFAR-10, n_train=40000, n_clients=1: 61.8\pm0.6
% Fashion-MNIST, n_train=100, n_clients=100: 63.9\pm0.6
% CIFAR-10, n_train=400, n_clients=100: 33.4\pm0.2

% Central
% 24000: 87.3\pm0.3
% 8000: 85.3\pm0.3
\begin{figure*}[t]
    \centering
    \subfloat[\oracle{}]{\label{heatmap-oracle}\includegraphics[width=0.315\textwidth]{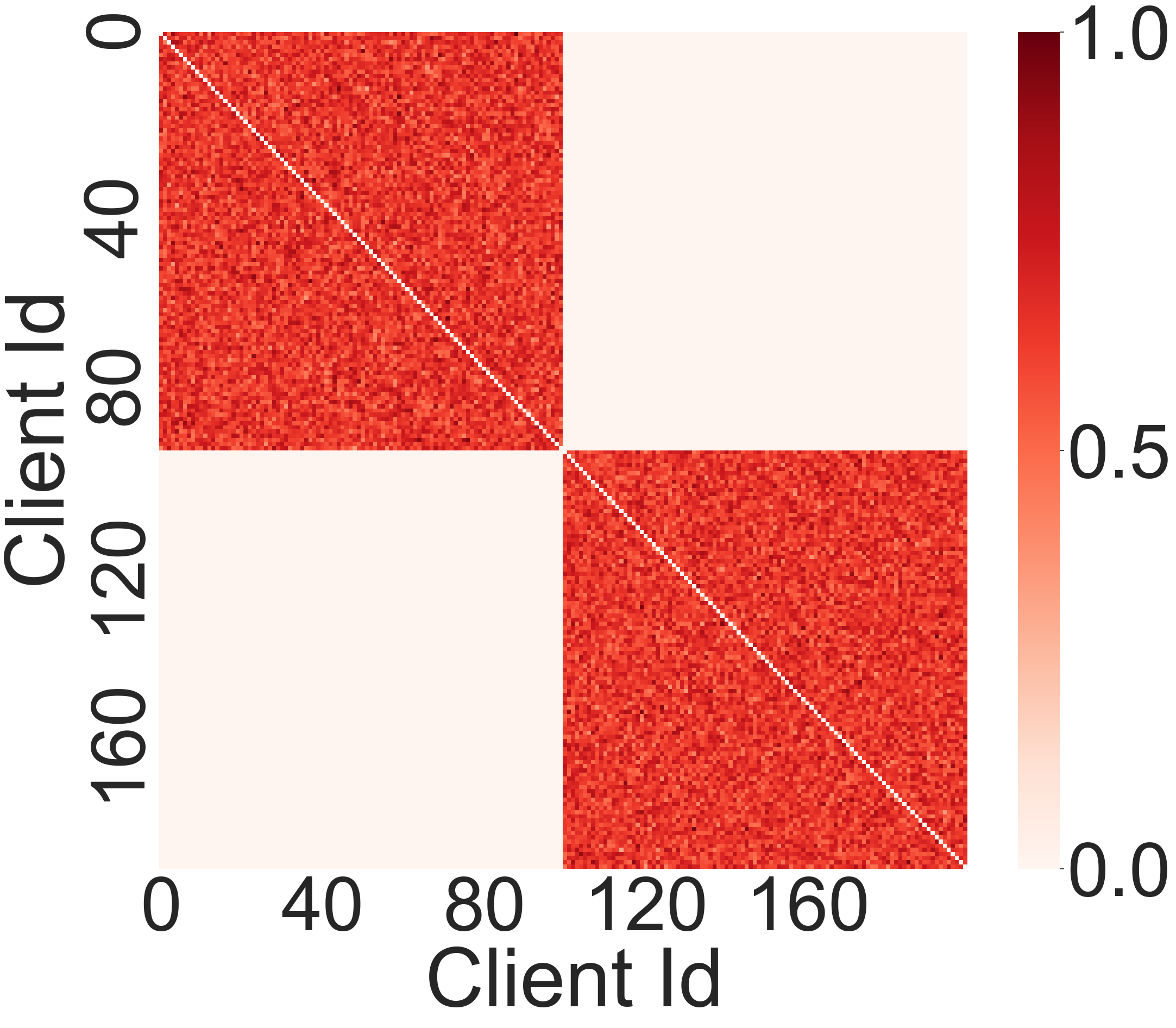}}\hfill
    \subfloat[\pens{}]{\label{heatmap-edu}\includegraphics[width=0.315\textwidth]{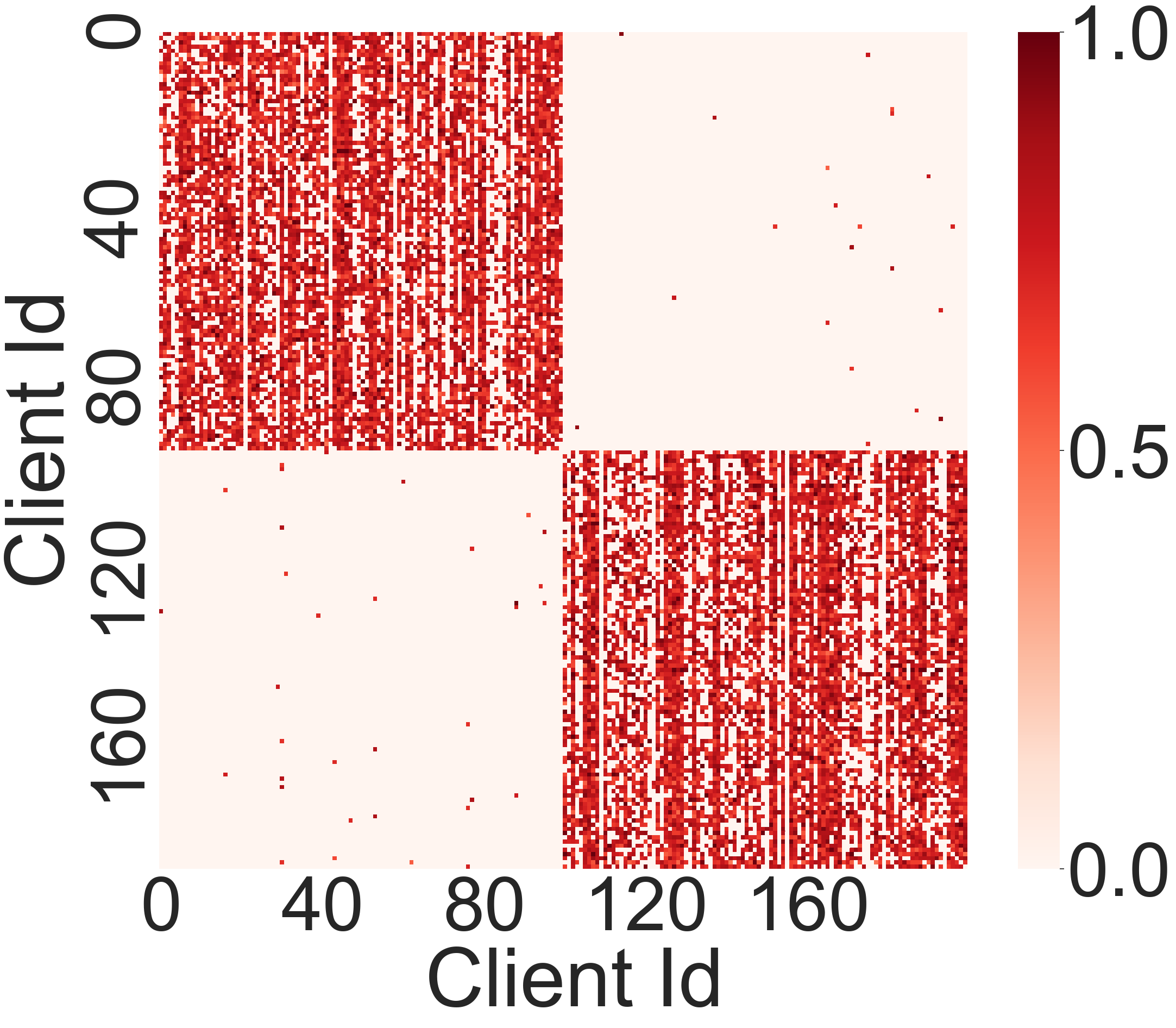}}\hfill
    \subfloat[Random]{\label{heatmap-eg0.5}\includegraphics[width=0.315\textwidth]{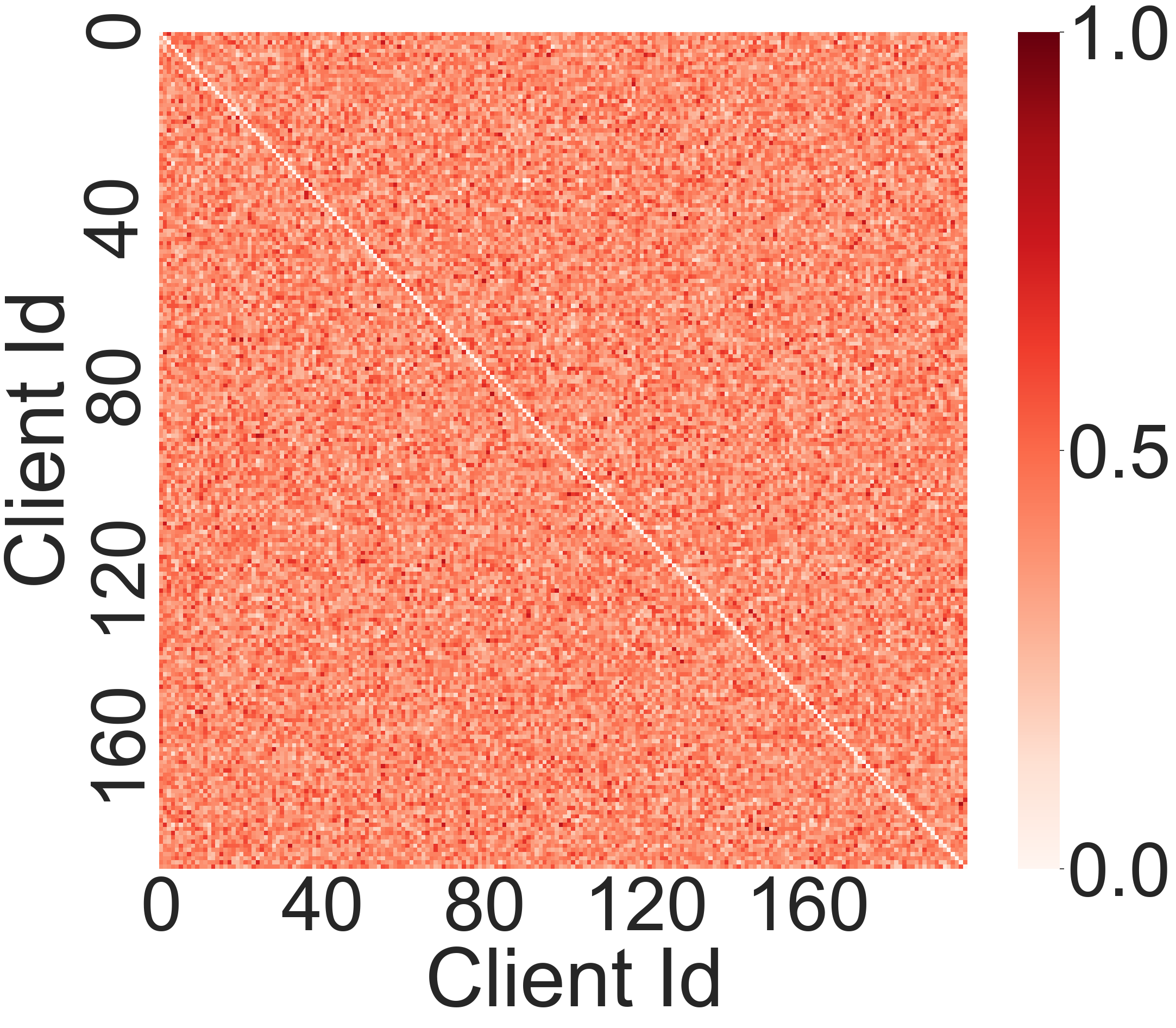}}\hfill
    \caption{Heatmaps over the communication pattern between clients on CIFAR-10 with two rotated $\mathcal{D}^r$ for (a) \oracle{} (b) \pens{} and (c) \texttt{random}. The reported values are normalized counts: a value close to 1 means that client $j$ received a model from client $i$ frequently. The clients are sorted so that clients with ID 0-99 have data distributions $\mathcal{D}^0(x)$ and clients with ID 100-199 have $\mathcal{D}^{180}(x)$.}
    \label{heatmaps}
\end{figure*}

% Värden för stora tabellen (varying size of dataset). Motsvarar figur \ref{datapoints}. 

%METHOD-SIZE OF DATASET: ACCURACY
%Local-400: 31.1\pm0.6
%Central-500.0: 65.9\pm0.8
%Central-400.0: 63.8\pm0.7
%Oracle-400: 55.5\pm1.0
%Oracle-200: 45.2\pm1.0
%Central-200.0: 58.3\pm0.5
%Oracle-500: 57.7\pm0.6
%PENS-200: 42.1\pm0.8
%PENS-300: 50.8\pm1.1
%Random-200: 41.6\pm1.9
%Oracle-300: 51.7\pm0.2
%Random-400: 52.0\pm0.9
%Local-300: 28.0\pm0.3
%Local-500: 33.3\pm0.7
%Random-300: 48.9\pm1.2
%Random-500: 54.1\pm0.8
%PENS-500: 57.2\pm1.3
%Local-200: 23.1\pm0.4
%PENS-400: 53.9\pm0.7
%Central-300.0: 60.5\pm1.1

% Värden för stora tabellen (varying number of clients). Motsvarar figur \ref{clients}. I central så motsvarar antalet clienter det antal clienter som datan har samlats in från.

%METHOD-NUMBER OF CLIENTS: ACCURACY
%Random-200: 48.4\pm0.8
%Random-50: 42.4\pm0.9
%Oracle-50: 45.0\pm2.2
%PENS-50: 42.5\pm1.5
%Local-50: 26.1\pm1.1
%Central-200.0: 65.9\pm0.8
%Random-100: 44.4\pm0.7
%Local-200: 26.0\pm0.9
%PENS-200: 50.8\pm0.8
%Oracle-100: 48.1\pm0.5
%Oracle-200: 53.0\pm0.8
%PENS-100: 47.8\pm0.2
%Central-50.0: 54.4\pm0.3
%Central-100.0: 60.0\pm0.9
%Local-100: 25.4\pm0.5

\section{Future work}
%Research on decentralized federated learning for deep neural networks has just begun. 
There are several interesting research directions left to explore which we had to limit ourselves from including in this paper. First, we assumed that all clients were able to communicate equally fast and at all times. This is a strong assumption that is not always true in many real world applications, and it would therefore be interesting to study \textbf{system heterogeneity} in a decentralized network where clients have different hardware and computational budgets. Second, we assumed that the decentralized \textbf{network topology} was fully connected. As future work, it would be interesting to study how \pens performs on other types of network topologies and how these effect the learning among clients.

\section{Conclusions}
In this work we have studied non-convex optimization of decentralized federated learning using deep neural networks in a non-iid data setting. Our experiments show that our proposed method \pens is able to efficiently aid clients to identify neighboring peers with similar data distributions in a fully decentralized FL setting and in that way guide the learning of the clients to achieve high performance in a non-iid data setting. \pens works by using the training loss to find clients in the network that share similar data distributions, which then focus on communicating with each other instead of random neighbors. Our results (figures \ref{datapoints_clients} and \ref{heatmaps}) suggest that given enough training data per client, \pens will reach the same accuracy as an oracle that is given \textit{perfect information} of the client data distributions. We have limited ourselves to the non-iid setting of covariate shift in this work. Meanwhile, we hypothesize that our proposed method also works well on other types of non-iid data, such as label distribution skew, concept shift or concept drift.

%\section*{Software and Data}

%If a paper is accepted, we strongly encourage the publication of software and data with the
%camera-ready version of the paper whenever appropriate. This can be
%done by including a URL in the camera-ready copy. However, \textbf{do not}
%include URLs that reveal your institution or identity in your
%submission for review. Instead, provide an anonymous URL or upload
%the material as ``Supplementary Material'' into the CMT reviewing
%system. Note that reviewers are not required to look at this material
%when writing their review.

%\section*{Acknowledgements}

%\textbf{Do not} include acknowledgements in the initial version of
%the paper submitted for blind review.

%If a paper is accepted, the final camera-ready version can (and
%probably should) include acknowledgements. In this case, please
%place such acknowledgements in an unnumbered section at the
%end of the paper. Typically, this will include thanks to reviewers
%who gave useful comments, to colleagues who contributed to the ideas,
%and to funding agencies and corporate sponsors that provided financial
%support.

\nocite{langley00}

\bibliography{example_paper}
\bibliographystyle{icml2021}

%appendix

\end{document}